\documentclass{llncs}
\usepackage{graphicx}
\usepackage{floatrow}
\usepackage{amsmath,amssymb,amsfonts, amstext}
\usepackage{longtable,booktabs}
\usepackage{array}
\usepackage{lscape} % for the big problem instances table
\usepackage{float}
\usepackage{multirow}
\usepackage{nicefrac} % for diagonal fractions
\usepackage{subcaption}
\usepackage{accents}

\begin{document}

\title{Shape-constrained Symbolic Regression with NSGA-III}
\titlerunning{Shape-constrained Symbolic Regression with NSGA-III}

\authorrunning{Christian Haider}

\author{
  Christian Haider
 }

\institute{
  Josef Ressel Center for Symbolic Regression\\
  Heuristic and Evolutionary Algorithms Laboratory\\
  University of Applied Sciences Upper Austria, Hagenberg, Austria\\
}

\maketitle % typeset the header of the contribution
\renewcommand{\thefootnote}{}
\footnotetext{\hspace{-0em}
Submitted manuscript to be published in \textit{Computer Aided Systems Theory - EUROCAST 2022: 18th International Conference, Las Palmas de Gran Canaria, Feb. 2022}.
}
\renewcommand\thefootnote{\arabic{footnote}}

\begin{abstract}
  Shape-constrained symbolic regression (SCSR) allows to include prior knowledge
  into data-based modeling. This inclusion allows to ensure that
  certain expected behavior is better reflected by the resulting models. The
  expected behavior is defined via constraints, which refer to the function
  form e.g. monotonicity, concavity, convexity or the models image boundaries.
  In addition to the advantage of obtaining more robust
  and reliable models due to defining constraints over the functions shape, the use of SCSR allows to find
  models which are more robust to noise and have a better extrapolation
  behavior. This paper presents a mutlicriterial approach to minimize the
  approximation error as well as the constraint violations. Explicitly the two
  algorithms NSGA-II and NSGA-III are implemented and compared against each
  other in terms of model quality and runtime. Both algorithms are
  capable of dealing with multiple objectives, whereas NSGA-II is a well
  established multi-objective approach performing well on instances with up-to 3
  objectives. NSGA-III is an extension of the NSGA-II algorithm and was
  developed to handle problems with "many" objectives (more than 3 objectives).
  Both algorithms are executed on a selected set of benchmark instances from
  physics textbooks. The results indicate that both algorithms are able to find
  largely feasible solutions and NSGA-III provides slight improvements in terms of
  model quality. Moreover, an improvement in runtime can be observed using the
  many-objective approach.

\end{abstract}

\section{Introduction}
Due to their high complexity, the behavior and certain phenomena of systems and
processes in critical application areas such as medical engineering or
financial systems cannot be feasibly modeled from first principles alone. Thus,
to handle the permanently increasing demand of high efficiency and accuracy as
well as scale of experiments, data-based modeling approaches have become a
standard.

The increase of computational power and the rising popularity of machine
learning algorithms for physical systems has given birth to a fourth paradigm of
science "(big) data-driven science". Traditionally, machine learning algorithms
are designed to train models, purely from data and achieve accurate predictions.
These often leads to models represented as highly complex functions forms (e.g.
neural networks) whose inner structure and working cannot be easily understood.
Such models are called black-box models. However, it must be ensured that models
found by data-driven algorithms conform to the functional principles of the
modeled system, to ensure correct functionality after deployment. This is a topic
especially relevant for scientific machine learning~\cite{baker2019workshop}.
For example, physical laws such as conservation of energy, thermodynamics, or
laws of gravity can be enforced for the modeling process when only a few data points are
given or when they are biased to prescribe some expected behavior. Since a
desired shape of the function form is enforced, these constraints are called
\emph{shape-constraints}~\cite{gupta2020multidimensional,ourPaper2,papp2014shape}.

A well-suited approach to incorporate shape constraints into modeling is
symbolic regression (SR)~\cite{koza1994genetic,poli2008field,koza2005genetic}.
Symbolic regression allows to find models represented as short closed-form
functions. This increases the interpretability and transparency ("Explainable
AI") of the models. SR finds the functional form as well as suitable parameters,
which distinguishes this approach from conventional regression methods, where
the functional form is already predefined and only the numerical coefficients
have to be adjusted and optimized (e.g. neural networks).

In~\cite{scsr} the authors proposed shape-constrained symbolic regression
(SCSR) a supervised machine learning approach that aims at both fitting
trainings data and compliance with the given shape-constraints. It is shown that
the resulting models, fit the training data well and even achieve slightly better
training errors than standard genetic programming (GP) models in some cases, and
that the models conform to the specified constraints.
%%Add reference to archive paper multi-objective approach

The paper is structured as follows: In section~\ref{sec:scr}, shape-constrained
regression is presented in detail as well as the use of a many-objective
approach. In section~\ref{sec:rw}, similar work in this area is highlighted and presented.
Section~\ref{sec:results} and~\ref{sec:sum} present the experiments performed
and results obtained, and conclude with a summary of the findings.

\section{Shape-constrained regression}
\label{sec:scr}
Shape-constrained regression (SCR) allows to enforce desired properties and
behavior by specifying constraints which refer to the shape of the function,
whereby the model is a function mapping real-valued inputs to real-valued outputs.
These constraints are formulated based on prior knowledge stemming from
empirical observations, domain experts knowledge, or physical princpiles.
Mathematically these constraints can be expressed through partial derivatives.
E.g. a monotonic increasing function can be enforced by a first order partial
derivative of a specific variable. Table~\ref{tab:constraints} highlights all
mathematical definition of constraints considered in this work.

\begin{table}[ht!]
  \center
  \caption{Mathematical formulation of constraints used in this work.}
  \label{tab:constraints}
  \centering
  \begin{tabular}{p{120pt}p{120pt}}
      Property & Mathematical definition\\
      \hline
      Positivity           & $f(X) > 0$ \\
      Negativity           & $f(x) < 0$ \\
      Model bounds         & $l \leq f(x) \leq u$ \\
      Monotonic-increasing & $\frac{\partial}{\partial x_{1}}f(x) > 0$ \\
      Monotonic-decreasing & $\frac{\partial}{\partial x_{1}}f(x) < 0$ \\
  \end{tabular}
\end{table}

The evaluation of the constraints often requires approximation methods due to
the need of finding the extrema of a non-linear function. For example a
monotonic increasing constraint - partial derivative w.r.t the variable needs to
be equal or greater than zero over the whole input domain. Therefore, the minimum
value of the partial derivative in the given domain has to be found. If the
model is non-linear, a non-linear optimization problem has to be solved, which
is often a NP-hard problem which results in the need of approximation methods.

Considering approximation methods, it can be distinguished between pessimistic
and optimistic approaches~\cite{optimisticPessimistic}. Pessimistic approaches
guarantee that the found solution is indeed valid but often discard valid
solution due to bound overestimation. Optimistic approaches can generate
false-positives but guarantee that solutions classified as infeasible are indeed
not feasible.

In this work interval arithmetic (IA) is used as a pessimistic approximation
method for constraint evaluation.

%%%%Formulation of shape-constrained regression%%%

\subsection{Many-objective approach}
In this work two multi/many-objective evolutionary search frameworks, as described
in~\cite{ourPaper2}, are used. Both presented algorithm start with a random
initialized population, represented as expression trees. Followed by a
repeating main-loop: fitness evaluation, parental selection, recombination, and
mutation.
Both algorithms are configured equally: max number of evaluation 500000,
tournament selection with a crowded group size of 5, single-subtree crossover,
and a mutator that either changes a single node or a subtree with a randomly
initialized subtree. Further parameter settings can be taken from~\cite{ourPaper2}.

The problem is modeled using a $1+n$ objectives approach - one main objective
and one objective for each constraint specified. The first objective is handled in
the data-based loss function and is minimizing the approximation error. In this
paper the error is calculated using the normalized mean squared error (NMSE) in percent, as
show in equation\ref{eq:nmse}, where $y$ represents the target vector and
$\hat{y}$ the prediction vector.

\begin{equation}
  \label{eq:nmse}
  \begin{aligned}
    NMSE(y,\hat{y}) = \frac{100}{var(y)N}\sum_{i=1}^{N}(y_{i}-\hat{y}_{i})^2
  \end{aligned}
\end{equation}

The following objectives are minimizing the constraint violations, whereas each
constraint is treated in a separate objective:

\begin{equation}
  \begin{aligned}
    P_{i} = P_{i}^{inf} + P_{i}^{sup}
  \end{aligned}
\end{equation}
where
\begin{equation}
  \begin{aligned}
    P_{i}^{inf} = |min(inf(f_{i}(x))-inf(c_{i}), 0)|\\
    P_{i}^{sup} = |max(sup(f_{i}(x))-sup(c_{i}),0)|
  \end{aligned}
\end{equation}

Where $sup(x)$ and $inf(x)$ are functions returning the superior and inferior
bounds of the intervals.%%!TODO more information

\section{Related work}
\label{sec:rw}
The most related article for this work is presented in~\cite{scsr}. The authors
introduce SCSR a method, which allows to use prior knowledge in a data-based
modelling approach. The authors presented a single objective approach, which
uses a feasible/infeasible population split. The proposed method was tested on a
set of instances taken from physics textbook. The results showed that using
a-priori knowledge helps with finding feasible solutions even on highly noise
data and on extrapolation.
Currently, the inclusion of additional domain knowledge in data-based modelling gets more and
more attention in literature and there are some recent paper targeting this
topic~\cite{baker2019workshop,rai2020explainable,liu2020certified}. Auguste et
al. presented two new methods to include monotonic constraints in regression and
classification trees~\cite{auguste}. In~\cite{kubalik2020symbolic} the authors
present a multi-objective symbolic regression approach to minimize the
approximation error on the training data as well as the constraint violations on
the constraint dataset. Therefore, they extended the NSGA-II algorithm and used
sampling to evaluate the constraints.
The inclusion of prior knowledge in data-based modeling also plays an
increasingly important role in modeling with neural networks as some recent
articles show~\cite{labelFreeNN,priorKnowledgeInNN,NNwithAsymptoticConstraints,liu2020certified,PINN}.

\section{Experiments and results}
\label{sec:results}
In this paper we compare the multi-objective algorithm NSGA-II to the
many-objective algorithm NSGA-III. The solution quality as well as the runtime
of both algorithms are used for comparison. The experiments are executed on a
set of different test instances.

\subsection{Problem instances}
To test both algorithms, expression from physics textbooks are used. Specifically,
instances from the \emph{Feynman Symbolic Regression
Database}~\cite{Feynman} are used. Therefore, only a subset of all instances are
used, selected by the reported difficulty in~\cite{Feynman} and only instances
where shape-constraints could be derived. To generate training data, 300
points are sampled uniformly at random out of the expressions from
table~\ref{tab:feynman}. Afterwards we split the points so that the first 10\%
and last 10\% are used as test set and the rest is used for training, this
represents a test set outside the hull of the training data. In
table~\ref{tab:constraintDefinitions} all derived constraints for each instance
are shown. The constraint's column shows in the first tuple the desired domain
constraint the following values define the constraints on each variable,
where $0$ means no constraint, $1$ defines a monotonic increasing
constraint, and $-1$ a monotonic decreasing function. The constraint definition
for the \emph{Pagie-1} instance follows the same principle, with the additional
restriction that the defined constraints are valid only in certain range of the
input variables, which is shown within the brackets.

\begin{table}[ht!]
  \center
  \caption{Problem instances taken from \cite{Feynman}}
  \label{tab:feynman}
  \begin{tabular}{p{70pt}p{120pt}}
      Instance & Expression\\
      \hline
      I.6.20        & $\operatorname{exp}\left( \frac{-{{\left( \frac{\theta}{\sigma}\right) }^{2}}}{2}\right) \frac{1}{\sqrt{2 \pi} \sigma}$\\
      I.9.18        & $\frac{G\, \mathit{m1}\, \mathit{m2}}{{{\left( \mathit{x2}-\mathit{x1}\right) }^{2}}+{{\left( \mathit{y2}-\mathit{y1}\right) }^{2}}+{{\left( \mathit{z2}-\mathit{z1}\right) }^{2}}}$\\
      I.30.5        & $\operatorname{asin}\left( \frac{\mathit{lambd}}{n d}\right)$\\
      I.32.17       & $\frac{1}{2} \epsilon c\, {{\mathit{Ef}}^{2}}\, \frac{8 \pi {{r}^{2}}}{3}\, \frac{{{\omega}^{4}}}{{{\left( {{\omega}^{2}}-{{{{\omega}_0}}^{2}}\right) }^{2}}}$\\
      I.41.16       & $\frac{h\, {{\omega}^{3}}}{{{\pi}^{2}}\, {{c}^{2}}\, \left( \operatorname{exp}\left( \frac{h \omega}{\mathit{kb} T}\right) -1\right) }$\\
      I.48.20       & $\frac{m\, {{c}^{2}}}{\sqrt{1-\frac{{{v}^{2}}}{{{c}^{2}}}}}$\\
      II.35.21      & ${n_{\mathit{rho}}} \mathit{mom} \operatorname{tanh}\left( \frac{\mathit{mom} B}{\mathit{kb} T}\right)$\\
      III.9.52      & $\frac{{p_d} \mathit{Ef} t}{h} {{\sin{\left( \frac{\left( \omega-{{\omega}_0}\right)  t}{2}\right) }}^{2}}$\\
      III.10.19     & $\mathit{mom}\, \sqrt{{{\mathit{Bx}}^{2}}+{{\mathit{By}}^{2}}+{{\mathit{Bz}}^{2}}}$\\
  \end{tabular}
\end{table}

Additionally, a regression problem from~\cite{pagie1997evoluationary} is added,
it is called Pagie-1 in the following.
It's a problem instance with two variables as follows:

\begin{equation}
  \begin{aligned}
    f(x,y)= \frac{1}{(1+x^{-4})}+\frac{1}{(1+y^-4)}
  \end{aligned}
\end{equation}
It is evaluated over the range of:
\begin{equation}
  \begin{aligned}
    -5 \leq x \leq 5 \quad \textrm{and} \quad -5 \leq y \leq 5 \quad (x,y \neq 0)
  \end{aligned}
\end{equation}

To generate trainings data for the Pagie-1 instance the training set consists of
data points spaced $0.4$ apart between the limits, which results with a set of 676
distinct $x,y$ data points.

\begin{table*}[ht!]
  \caption{Shape constraints used for each problem instance. \emph{Input space}
  column refers to the variable domains. \emph{Constraints} column represents
  the defined constraints over each variable. The first tuple represents the
  allowed model range the following values represent either no constraint over
  the variable for value 0, monotonic decreasing for value -1 or monotonic
  increasing for value 1.}
    \label{tab:constraintDefinitions}
    \resizebox{\textwidth}{!}{
    \begin{tabular}{p{1.7cm}>{$}l<{$}>{$}l<{$}}
        Instance & \text{Input space} & \text{Constraints}\\
        \hline
        I.6.20        & (\sigma, \theta) \in [1..3]^2 & ([0..\infty], 0, -1)\\
        I.9.18        &
        \left(\mathit{x1},\mathit{y1},\mathit{z1},\mathit{m1},\mathit{m2},G,\mathit{x2},\mathit{y2},\mathit{z2}\right) & ([0..\infty], -1, -1, -1, 1, 1, 1, 1, 1, 1)\\
        &\in [3..4]^3\times[1..2]^6&\\
        I.30.5        & \left( \mathit{lambd},n,d\right) \in [1..5]^2\times[2..5]& ([0..\infty], 1, -1, -1)\\
        I.32.17       & \left( \epsilon,c,\mathit{Ef},r,\omega,{{\omega}_0}\right) \in [1..2]^5\times[3..5]& ([0..\infty], 1, 1, 1, 1, 1, -1)\\
        I.41.16       & \left( \omega,T,h,\mathit{kb},c\right)  \in [1..5]^5& ([0..\infty], 0, 1, -1, 1, -1)\\
        I.48.20       & (m, v, c) \in [1..5]\times[1..2]\times[3..20]& ([0..\infty], 1, 1, 1)\\
        II.35.21      & \left( {n_{\mathit{rho}}},\mathit{mom},B,\mathit{kb},T\right)  \in [1..5]^5& ([0..\infty], 1, 1, 1, -1, -1)\\
        III.9.52      & \left( {p_d},\mathit{Ef},t,h,\omega,{{\omega}_0}\right) \in  [1..3]^4\times[1..5]^2 & ([0..\infty], 1, 1, 0, -1, 0, 0)\\
        III.10.19     & \left( \mathit{mom},\mathit{Bx},\mathit{By},\mathit{Bz}\right)  \in [1..5]^4 & ([0..\infty], 1, 1, 1, 1)\\
        Pagie-1       & \left( x,y\right) \in [-5..5]^2 & ([0..2],-1(x<0),1(x>0),-1(y<0),1(y>0))\\
    \end{tabular}}
\end{table*}

\subsection{Results}
Table~\ref{tab:resultsTest} shows the result of 10 independent runs over all test
instances. The error is represented as normalized mean squared error (NMSE) in
percent. The left most column shows the instance and in the following two
columns the results of both algorithms NSGA-II and NSGA-III are shown. It  can
be observed that both algorithms give similar results, but NSGA-III gives
slightly better results over all instances, but without being statistically
significant. In table~\ref{tab:runtime} the runtime performance of both
algorithms is compared. The left most columns shows again the instance followed
by the two columns showing the median runtime measured in seconds. The results
are similar to the error comparison in table~\ref{tab:resultsTest}, whereas the
NSGA-III has advantage over all instances.
\begin{table*}[!ht]
  \caption{Median test error (NMSE in \%)}
  \centering
  \label{tab:resultsTest}
  \begin{tabular}[]{c@{}l>{$}r<{$}>{$}r<{$}>{$}r<{$}}
      & & \text{NSGA-II} & \text{NSGA-III}\\
      \hline
      & I.6.20       & 20.88  & \textbf{19.14}\\
      & I.30.5       & 7.32   & \textbf{6.24}\\
      & I.32.17       & 7.17   & \textbf{6.38}\\
      & I.41.16      & 18.50  & \textbf{15.21}\\
      & I.48.20      & 24.19 & \textbf{22.58}\\
      & II.35.21     & 14.60  & \textbf{14.54}\\
      & III.9.52     & 89.03  & \textbf{89.00}\\
      & III.10.19    & 11.30  & \textbf{10.62}\\
      & Pagie-1      & 46.41  & \textbf{40.71}\\
\hline
  \end{tabular}
\end{table*}

\begin{table*}[!ht]
  \caption{Median runtime (in seconds)}
  \centering
  \label{tab:runtime}
  \begin{tabular}[]{c@{}l>{$}r<{$}>{$}r<{$}>{$}r<{$}}
      & & \text{NSGA-II} & \text{NSGA-III}\\
      \hline
      & I.6.20       &  1798.02 & 1407.67 \\
      & I.30.5       &  3621.99 & 3604.91 \\
      & I.32.17      &  5812.10 & 4504.23 \\
      & I.41.16      &  3858.61 & 2879.05 \\
      & I.48.20      &  2825.38 & 1647.43 \\
      & II.35.21     &  3217.67 & 3045.41 \\
      & III.9.52     &  3009.62 & 2064.16 \\
      & III.10.19    &  3939.14 & 2254.29 \\
      & Pagie-1      &  4800.77 & 4105.86\\
\hline
  \end{tabular}
\end{table*}

\section{Summary}
\label{sec:sum}
In this paper two multi-objective methods for shape-constrained symbolic
regression are evaluated and compared to each other. The comparison of both
methods was mainly motivated by the article~\cite{} in which the advantages of
many-objective algorithms over multi-objective algorithms, when having more than
3 objectives is shown.

The two algorithms have been integrated in the heuristic and evolutionary
algorithm framework (HeuristicLab) and are benchmarked on a set of equation from
physics textbooks \emph{Feynman Symbolic Regression Database}.

The results showed that using many-objective algorithms over multi-objective
algorithms has slightly advantages in some instances but for a significant
difference the instances had too less objectives. It is also shown that using the
many-objective approach can help with runtime.

Although no statistical significant differences were achieved, it is shown
that many-objective algorithms can help with performance increases on instances
with more than 3 objectives. A more detailed comparison on instances with more
than 10 objectives will be investigated in further research.

%%Überleitung zu Multi-objective

\subsection*{Acknowledgement}
The financial support by the Christian Doppler Research Association, the
Austrian Federal Ministry for Digital and Economic Affairs and the National
Foundation for Research, Technology and Development is gratefully acknowledged.

\bibliographystyle{splncs}
\bibliography{bib}

\end{document}